\documentclass{article}
\usepackage{amssymb}

\usepackage[acronym]{glossaries}
\usepackage{booktabs}
\usepackage{multirow}
\usepackage{comment}
\usepackage{graphicx}
\usepackage{subcaption}
\usepackage{bm}
\usepackage{amsmath}
\usepackage{amssymb}
\usepackage{amsfonts}
\usepackage{amstext}
\usepackage{amsthm}
\usepackage{enumitem}
\usepackage{url}
\usepackage{array}

\usepackage[final]{corl_2025} 

\title{KDPE: A Kernel Density Estimation Strategy for Diffusion Policy Trajectory Selection}

\author{
  Andrea Rosasco$^{1,2}$ \And Federico Ceola$^1$ \And Giulia Pasquale$^1$ \And Lorenzo Natale$^1$\\
  \Inst$^1$Istituto Italiano di Tecnologia, $^2$University of Genoa \\
  \texttt{\{andrea.rosasco, federico.ceola, giulia.pasquale, lorenzo.natale\}@iit.it}
}

\begin{document}
\maketitle

\begin{abstract}
Learning robot policies that capture multimodality in the training data has been a long-standing open challenge for behavior cloning. Recent approaches tackle the problem by modeling the conditional action distribution with generative models. One of these approaches is Diffusion Policy, which relies on a diffusion model to denoise random points into robot action trajectories. While achieving state-of-the-art performance, it has two main drawbacks that may lead the robot out of the data distribution during policy execution. 
First, the stochasticity of the denoising process can highly impact on the quality of generated trajectory of actions. Second, being a supervised learning approach, it can learn data outliers from the dataset used for training.  
Recent work focuses on mitigating these limitations by combining Diffusion Policy either with large-scale training or with classical behavior cloning algorithms. Instead, we propose KDPE, a Kernel Density Estimation-based strategy that filters out potentially harmful trajectories output of Diffusion Policy while keeping a low test-time computational overhead.
For Kernel Density Estimation, we propose a manifold-aware kernel to model a probability density function for actions composed of end-effector Cartesian position, orientation, and gripper state.  
KDPE overall achieves better performance than Diffusion Policy on simulated single-arm tasks and real robot experiments.

Additional material and code are available on our project page: \url{https://hsp-iit.github.io/KDPE/}.
\end{abstract}



\section{Introduction}
Diffusion Policy (DP)~\citep{chi2023diffusionpolicy} has recently emerged as a powerful robotic policy representation due to its capability of handling multimodal behaviors. DP models the robot policy as a Denoising Diffusion Probabilistic Model (DDPM)~\citep{ho2020denoising}, and during policy execution denoises a set of randomly sampled trajectory points into the trajectory that the robot is requested to execute with receding horizon control~\citep{mayne1988receding}. While enabling the robot to capture multimodality in the demonstrations used for training, the choice of the random points to initialize the denoising process performed at inference can lead the robot to execute different trajectories, given the same observation. This may be problematic if the sampled trajectory is an outlier with respect to modes that are most represented in the training data, leading the robot into a state which may be out of the distribution of the demonstrations dataset.

To overcome this limitation, we propose KDPE: a strategy to sample trajectories that are representative of the modes learned by the policy. We propose to compute a set of  \( N \) trajectories, by performing in parallel  \( N \) denoising processes from  \( N \) different starting noise samples conditioned on the same current observation. We then estimate the Probability Density Function (PDF) on the last actions in the trajectories generated by Diffusion Policy, using Kernel Density Estimation (KDE), and select the trajectory associated to the action with the highest density. 

The actions we consider are composed of end-effector pose and gripper state. Modeling a KDE on a population of actions requires a kernel function that relates end-effector position, orientation and gripper state. While probability distributions on end-effector poses have previously been proposed \citep{barfoot2014, igor2014}, none of them integrate the gripper state. We modify the probability distribution introduced in \citep{barfoot2014} to handle the gripper state component. 

In principle, the multimodality of the trajectories predicted by DP can be modeled with non-parametric clustering algorithms~\citep{1000236}. However, these approaches require non-negligible convergence time when applied for outlier rejection~\citep{dbscan} in closed-loop control tasks. KDPE, instead, manages to filter and sample among a subset of the trajectories predicted by DP with a significantly lower computational overhead, making it suitable for the target visuomotor closed-loop control tasks. While KDPE uses DP for trajectory generation, the core idea of sampling multiple trajectories and filtering them through KDE can be applied to any probabilistic generative policy.

We evaluate KDPE against DP on the four RoboMimic~\citep{mandlekar2022matters} single-arm tasks used in~\citep{chi2023diffusionpolicy} to benchmark DP, and on three tasks from the MimicGen benchmark~\citep{mandlekar2023mimicgen}. We train DP on these tasks, and evaluate the models with KDPE, achieving an overall improvement in terms of average success rate. We then evaluate KDPE's robustness to visual perturbations by changing the color of an object in the task environments and show that KDPE maintains performance closer to the non-shifted setting than DP.

Finally, using a Franka Emika Panda~\citep{9721535} manipulator, we test KDPE on three real-world tasks: \textit{PickPlush}, a tabletop picking task where a plush has to be grasped and lifted by the robot, \textit{CubeSort}, a multi-step multimodal task where the robot has to pick two cubes and place them in the cup of the corresponding color, and \textit{CoffeeMaking}, a fine-grained multi-step task where the robot has to load a coffee machine with a pod. Overall, KDPE outperforms DP in terms of success rate, and qualitatively shows smoother behavior. 

In summary, the contributions of the paper are:
\begin{itemize}
    \item \textbf{KDPE}: a KDE-based approach for selection of DP action trajectories that are representative of the different action modalities.
    \item A \textbf{manifold-aware kernel for KDE} on robotic actions composed of Euclidean end-effector position and gripper aperture, and non-Euclidean end-effector orientation.
    \item An \textbf{extended quantitative evaluation of KDPE} on seven simulated  tasks from two different benchmarks and on three real-world tasks.
    \item A \textbf{trajectory visualizer} for qualitative analysis of manipulation policies.
\end{itemize}

\section{Related Work}

There is a huge effort in the robotics community to create generalist robot policies~\citep{octo_2023, kim24openvla, bousmalis2024robocat} from large-scale datasets~\citep{open_x_embodiment_rt_x_2023, khazatsky2024droid}, either with large transformer-based models~\citep{brohan2022rt, bousmalis2024robocat, octo_2023, black2024pi_0, intelligence2025pi_}, or parameterizing robot policies as language tokens~\citep{pmlr-v229-zitkovich23a, kim24openvla, team2025gemini}. However, for single tasks, there is evidence~\citep{kim24openvla} that DP~\citep{chi2023diffusionpolicy} achieves better performance than large generalist models~\citep{octo_2023, kim24openvla}.

DP~\citep{chi2023diffusionpolicy}, together with its flow matching variant~\citep{zhang2023bootstrap}, is currently one of the most widely used approaches for behavior cloning for robotic manipulation tasks. Its policy parameterization as a Denoising Diffusion Probabilistic Model (DDPM) allows to model the multimodality in the training data. Several recent works extend the original DP problem formulation to learn goal-conditioned~\citep{reuss2023goal}, or language-conditioned~\citep{ha2023scaling, dasari2024ditpi} tasks. Octo~\citep{octo_2023}, instead, uses DP to parameterize a generalist robot manipulation policy on top of a large transformer-based modular backbone for transferability across tasks and embodiments. Approaches like the one presented in~\citep{zhang2024diffusion} combine DP with more classical behavior cloning algorithms like DAgger~\citep{ross2011reduction} to reduce the compounding errors typical of pure imitation learning algorithms.

Recent work has explored failure detection and mitigation strategies to enhance the reliability of robot policies. Sentinel~\citep{sentinel} identifies potential failures by analyzing the consistency of overlapping trajectory segments sampled at different time-steps, and leverages vision-language models to detect misalignment between actions and visual context. While effective, such methods often detect failures after the policy has already entered an out-of-distribution state, making recovery challenging. RoboMD~\citep{RoboMD} addresses robustness by systematically identifying failure modes through DRL-guided exploration across diverse environmental conditions. RoboMD fine-tunes policies on these identified failure modes. This \textit{train-deploy-analyze-retrain} pipeline, however, can be impractical for real-world deployment. 
V-GPS \cite{nakamoto2024steering} improves robot policies at inference time by using a value function network to rank actions. However, this requires to train a value function network with offline Reinforcement Learning and design a reward function.
In contrast, KDPE operates at inference time without requiring any additional training. By leveraging the inherent diversity of DP’s trajectory sampling process, KDPE mitigates failures proactively, selecting trajectories that align with the statistical modes of the learned distribution, rather than relying on post-hoc detection or costly retraining. 

A growing body of literature in Natural Language Processing studies \textit{inference or test-time scaling} methods that select one among multiple output samples generated (e.g. by a Large Language Model) to improve the performance of the trained model~\citep{wu2024scaling,brown2024large, snell2024scaling}.
Furthermore, there is recent research interest in the application of inference-time scaling to, e.g., text-to-image diffusion models~\citep{ma2025inferencetimescalingdiffusionmodels, li2025reflectditinferencetimescalingtexttoimage}. In~\citep{xie2025sana} the effectiveness of the easiest \textit{best-of-N sampling} (generation of a high number of random samples and selection of the best one by using an external evaluator) is proved experimentally. To the best of our knowledge, however, KDPE is the first method that improves DP at inference time based on statistical properties of the trajectories distribution.

\section{Methodology}
\label{sec:methodology}

\subsection{Diffusion Policy}

DP adapts DDPM, a generative diffusion model which is commonly used for image generation, to behavior cloning for robotic manipulation tasks in a receding horizon fashion. DP, given an observation of the environment, outputs action trajectories of shape $T \times D$, where $T$ is the number of time-steps and $D$ is the action dimensionality. These trajectories are generated through a denoising process that progressively refines Gaussian noise by iteratively subtracting a learned gradient field. This process allows the policy to model multimodal trajectory distributions and maintain the multimodality of task demonstrations. The denoising process to generate a trajectory is defined as:
\begin{equation}
    \mathbf{A}^{k-1} = \alpha(\mathbf{A}^k - \gamma \epsilon_{\theta}(\mathbf{A}^k, k) + \mathcal{N}(0, \sigma^2 I)),
\end{equation}
where $\epsilon_{\theta}$ is the noise prediction network with parameters $\theta$ and $\mathbf{A}^k$ is a noisy sample going through a denoising step. This process is repeated for $K$ steps, starting with $\mathbf{A}^K$ as randomly sampled Gaussian noise, to output the trajectory $\mathbf{A}^0$. 
During training, a trajectory is sampled from the dataset and is perturbed by adding the appropriate amount of noise corresponding to a random denoising step $k$. The noise prediction network $\epsilon_{\theta}$ takes as input the noisy sample and is optimized to predict the noise $\epsilon^k$ that has been added to the ground-truth trajectory $\mathbf{A}^0$ with the following loss function:
\begin{equation}
\mathcal{L} = MSE(\epsilon^k, \epsilon_{\theta}(\mathbf{A}^0 + \epsilon^k, k)).
\end{equation}

\subsection{KDPE}
\label{sec:kdpe}
KDPE enhances Diffusion Policy by scoring the population of trajectories via Kernel Density Estimation (KDE)~\citep{parzen1962estimation} and selecting the best using \textit{best-of-N sampling}. Our approach combines a manifold-sensitive kernel with multi-hypothesis sampling from the diffusion process. Given an observation $\mathbf{o}_t$, we sample $N$ independent and identically distributed action trajectories $\{\mathbf{A}_i\}_{i=1}^N \sim p_{\boldsymbol{\theta}}(\mathbf{A}|\mathbf{o}_t)$, where $\mathbf{A}_i \in \mathbb{R}^{T \times D}$, and estimate the PDF of the last action of the trajectories, hereinafter defined as $\mathbf{a}_i\in \mathbb{R}^{D}$,  via KDE. This allows to obtain the probability density of every action and use it to discard outlier trajectories.

To model the PDF, KDE requires a unified kernel over all the action components representing end-effector position and orientation, and gripper aperture. While the multivariate Gaussian kernel is a natural choice for Euclidean domains, it cannot directly handle data lying on different manifolds such as rotations in $\mathrm{SO}(3)$, being a non-Euclidean space. 

\begin{figure}[t]
    \centering
    \includegraphics[width=1\textwidth]{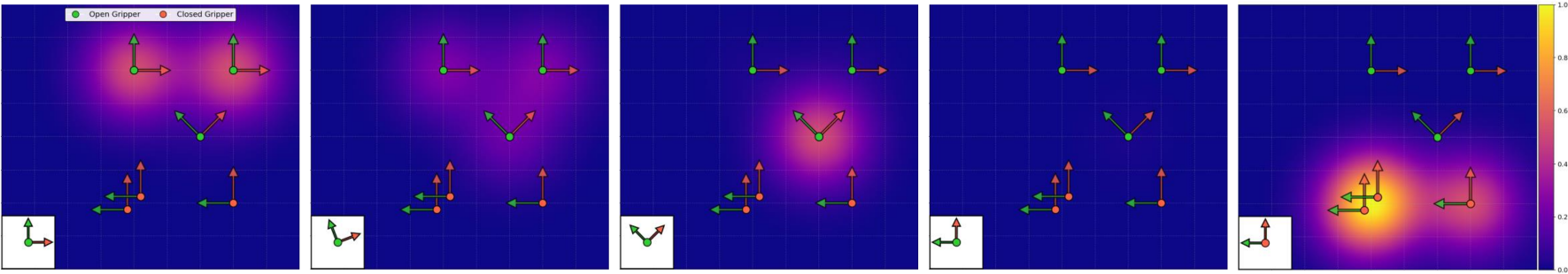}
    \caption{
    Visualization of the PDF estimated via KDE with the proposed manifold-aware kernel. We perform KDE on a population of 6 planar end-effector actions represented as reference frames in the plots. Three of them represent open grippers (green circle), while the other three represent closed grippers (red circle). The color of each point in the heatmaps represents the density value of an action at the corresponding 2D location, that has orientation and gripper state showed by the indicator frame in the white square of each plot. From left to right we vary the rotation of the indicator frame from 0 to 90 degrees and observe that the densities returned by KDE at different locations vary accordingly, spiking when the probed actions are close to the ones used for PDF modeling. The plots show how KDE correctly handles multimodality by providing the highest density values for the most well represented samples. The two rightmost plots show how the gripper state is correctly handled by the KDE.}
    \label{fig:kde-viz}
    \vspace{-0.2cm}
\end{figure}

\noindent \textbf{Manifold-Aware Kernel Density Estimation}
We propose a modification of the probability distribution presented in \citep{barfoot2014}  to handle the pose of the end-effector together with the gripper state.  Although DP outputs orientation in the 6D matrix representation introduced in \citep{zhou2019rotations6d}, for the sake of clarity, we consider their matrix representation in $\mathrm{SO}(3)$. Our kernel function is defined by the equation
\begin{equation}
k(\mathbf{a}_i,\mathbf{a}_j) = \frac{1}{\sqrt{(2\pi)^D|\mathbf{H}|}} \exp\left(-\frac{1}{2}\mathbf{\Delta}_{ij}^\top \mathbf{H}^{-1} \mathbf{\Delta}_{ij}\right),
\end{equation}
where $\mathbf{a}_i = [\mathbf{t}_i; \mathbf{R}_i; \mathbf{g}_i]$ and $\mathbf{a}_j = [\mathbf{t}_j; \mathbf{R}_j; \mathbf{g}_j]$ are actions composed of position, rotation, and gripper aperture components. $\mathbf{H}$ is the covariance matrix defined as $\mathbf{H} = \text{diag}\big(\sigma_{\text{pos}}^2 \mathbf{I}_3, \, \sigma_{\text{rot}}^2 \mathbf{I}_3, \, \sigma_{\text{grip}}^2 \mathbf{I}_1\big)$, and $\mathbf{\Delta}_{ij} \in \mathbb{R}^D$ represents the difference between manifold-specific components defined as:
\begin{equation}
\mathbf{\Delta}_{ij} = [\underbrace{\mathbf{t}_i - \mathbf{t}_j}_{\text{Euclidean}}; 
\underbrace{\log(R_j^{T}R_i)^\vee}_{\mathrm{SO}(3)}; 
\underbrace{\mathbf{g}_i - \mathbf{g}_j}_{\text{Euclidean}}]
\end{equation}
For position and gripper components, standard Euclidean differences suffice. For rotations represented as rotation matrices $R_i, R_j \in \mathrm{SO}(3)$, we compute the transformation  from $R_j$ to $R_i$  and convert it to its axis-angle representation using the Lie group logarithm map $\log: \mathrm{SO}(3) \to \mathfrak{so}(3)$ followed by the vee operator $(\cdot)^\vee: \mathfrak{so}(3) \to \mathbb{R}^3$ to obtain a representation in the tangent space. 
From the definitions above, we can rewrite $ -\frac{1}{2}\mathbf{\Delta}_{ij}^\top \mathbf{H}^{-1} \mathbf{\Delta}_{ij} $ as 
\begin{equation}
   -\frac{1}{2}\left(
\frac{\|\mathbf{t}_i - \mathbf{t}_j\|^2}{\sigma_{\text{pos}}^2} + 
\frac{\|\log(R_j^\top R_i)^\vee\|^2}{\sigma_{\text{rot}}^2} + 
\frac{(\mathbf{g}_i - \mathbf{g}_j)^2}{\sigma_{\text{grip}}^2}
\right),
\end{equation}
where $\|\log(R_j^\top R_i)^\vee\|^2$ is the geodesic distance representing  the minimal rotation angle $\theta_{ij}$ required to align $R_i$ and $R_j$:
\[
d_{SO(3)}(R_i, R_j)  = \|\log(R_j^\top R_i)^\vee\|^2 = \|\boldsymbol{\theta_{ij}}\| = \theta_{ij}.
\]
In Fig.~\ref{fig:kde-viz}, we show how the proposed manifold-aware kernel allows to model PDFs for actions comprising end-effector position and orientation, and gripper aperture. 

\noindent \textbf{Density Estimation and Action Selection}
We model the PDF underlying the $N$ actions $\{\mathbf{a}_i\}_{i=1}^N$ predicted by DP with KDE, by combining kernel densities across all  actions. Specifically, we compute the density of a generic action $\tilde{\mathbf{a}}$ as:
\begin{equation}
\rho(\tilde{\mathbf{a}}) = \frac{1}{N}\sum_{i=1}^N k(\tilde{\mathbf{a}}, \mathbf{a}_i)
\end{equation}
\noindent Finally, we compute densities $\rho_i = \rho(\mathbf{a}_i)$ for each of the $N$ final actions predicted by DP and we select the trajectory $\mathbf{A}_i$ corresponding to the action $\mathbf{a}_i$ with the highest ${\rho}_i$.

\section{Experimental Setup}

\subsection{Simulated Environments}
\label{sec:expsetup_sim}

\begin{figure*}[t!]
    \centering
    \includegraphics[width=1.0\linewidth]{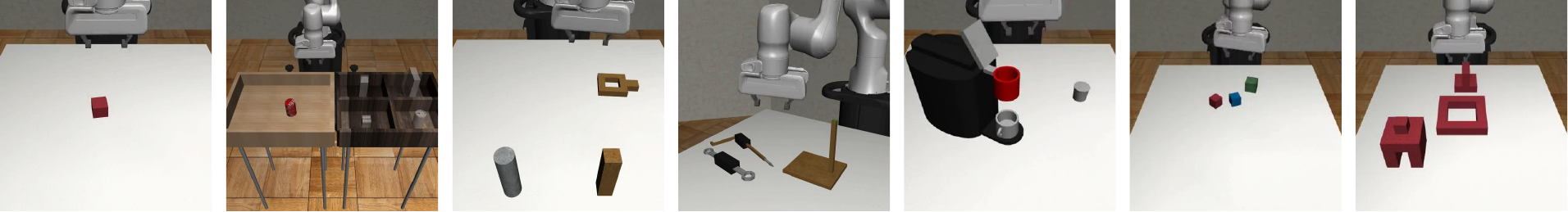}
    \caption{RoboMimic (\textit{Lift}, \textit{Can}, \textit{Square} and  \textit{ToolHang}) and MimicGen (\textit{Coffee}, \textit{Stack} and \textit{Assembly}) tasks considered for KDPE's evaluation.}
    \label{Fig:benchmark}
    \vspace{-0.5cm}
\end{figure*}

We evaluate KDPE on seven tasks from two different benchmarks: RoboMimic~\citep{robomimic2021} and \mbox{MimicGen}~\citep{mandlekar2023mimicgen}. We consider four RoboMimic tasks: \textit{Lift}, \textit{Can}, \textit{Square}, and \textit{ToolHang} (illustrated in Fig.~\ref{Fig:benchmark}). This subset of RoboMimic tasks corresponds to the single-arm image-based tasks used to evaluate DP in~\citep{chi2023diffusionpolicy}. RoboMimic provides two datasets for \textit{Lift}, \textit{Can}, and \textit{Square}: one collected by proficient human operators (\textit{ph}) and another by a mix of proficient and non-proficient operators (\textit{mh}), resulting in lower data quality for \textit{mh}. For \textit{ToolHang}, only the \textit{ph} dataset is available. We also test KDPE on three MimicGen tasks: \textit{Coffee}, \textit{Stack Three} (\textit{Stack} for conciseness) and \textit{Three Piece Assembly} (\textit{Assembly}). We chose this benchmark because it shares the same structure as RoboMimic, making the integration with the evaluation framework seamless and allowing us to test KDPE on additional challenging tasks. 

To assess the robustness of KDPE under visual domain shift, we compare its performance with DP on perturbed variations of the original environments where we slightly modify the color of an object. Specifically, we remove the original texture from specific objects and set their new color as the average value of the original texture with a decreased 10\% lightness value under the HSL color scheme. The specific objects modified for this experiments are shown in App.~\ref{app:perturbations}.

\subsection{Comparison with Diffusion Policy}
\label{ssec:benchmark-setup}
To compare KDPE with Diffusion Policy (DP), we train DP for $80k$ training steps. For each experimental setting, we rollout the methods for 100 random resets of the environment. The random sequence of environment initializations and the DDPM noise schedules are kept fixed across experiments.
During policy rollout, we sample a population of 100 trajectories at every inference step, and one trajectory is selected differently for each method. The DP baseline uniformly samples the output trajectory from the population. For KDPE, we perform KDE on the eighth actions of the population of trajectories, as eight is the action execution horizon used in DP. If the selection method is non-deterministic, as it is the case for DP, we run the full set of rollouts three times and report the average result.

Additionally to the comparison with DP, we report performance on two modified versions of KDPE: KDPE-OOD and Tr-KDPE.
KDPE-OOD is a modification of KDPE, where we choose the trajectory associated to the least represented action, i.e., the action with the minimum density $\{\rho_i\}_{i=1}^N$ (see Sec.~\ref{sec:kdpe}). We evaluate this method to further support the need for outliers rejection in the output trajectories of DP. Tr-KDPE, instead, is a modification of KDPE which uses conditional KDE and a first order Markovian assumption to estimate the PDF of the population of whole trajectories. We report the details of Tr-KDPE in App.~\ref{app:trkdpe}.

We present results for both the CNN-based (DP-C, KDPE-C, KDPE-OOD-C, Tr-KDPE-C) and the Transformer-based (DP-T, KDPE-T, KDPE-OOD-T, Tr-KDPE-T) models. For all methods, we use the same hyperparameters, e.g. KDE kernel bandwidths, reported in App.~\ref{app:params}.

\subsection{Real Robot}
\label{ssec:robot-setup}

\begin{figure*}[t!]
    \centering
    \includegraphics[width=\textwidth]{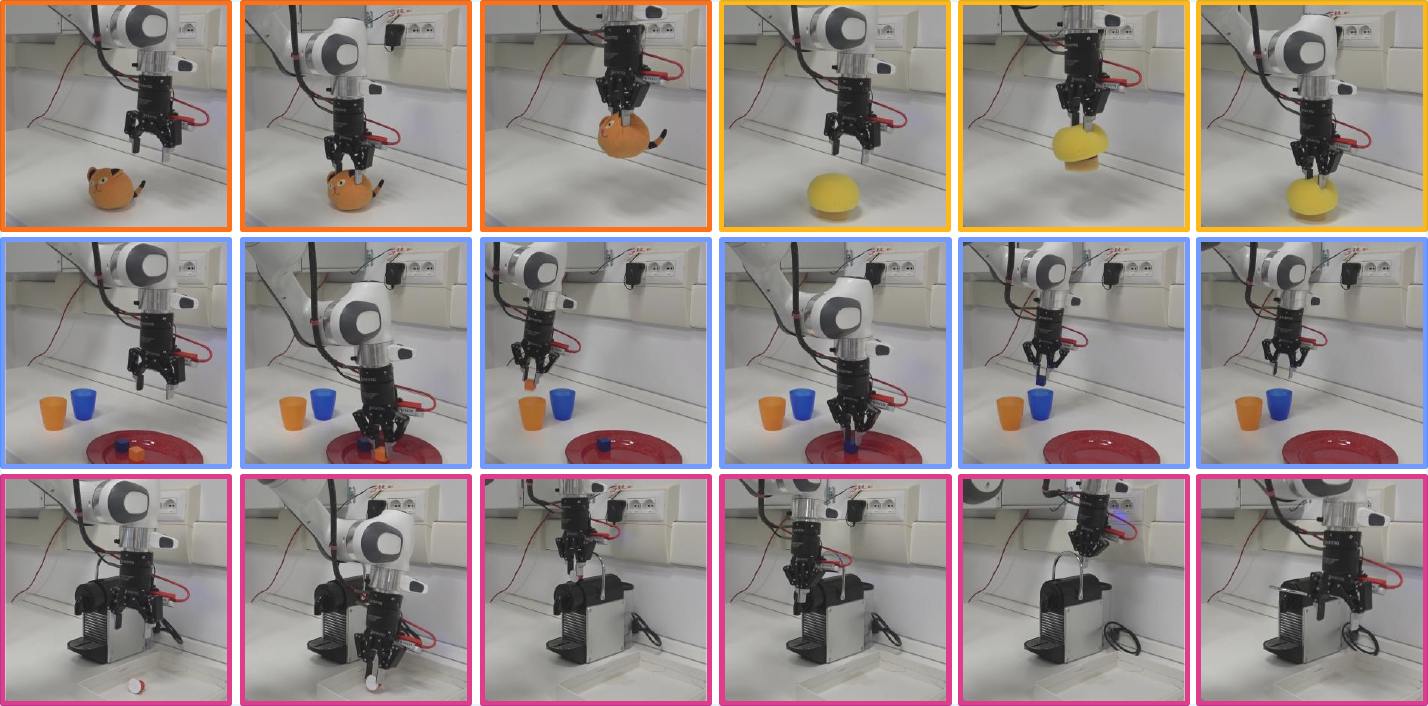}
    \caption{KDPE autonomously executing the real-robot tasks: \textit{PickPlush} (orange border) and its variant \textit{PickSponge} (yellow), \textit{CubeSort} (blue) and \textit{CoffeeMaking} (purple). }
    \label{Fig:real-tasks}
    \vspace{-.5cm}
\end{figure*}

We collect datasets to train the policies by using a \textit{Meta Quest 3} VR headset to teleoperate a \textit{Franka Emika Panda} robotic arm equipped with a \textit{Robotiq 2f-85} gripper. We provide visual RGB observations to the robot from an \textit{Intel(R) RealSense D415} mounted on the wrist of the robot and an external \textit{Intel(R) RealSense D405} (Fig. \ref{Fig:real-tasks}). We collect $50$ demonstrations for \textit{PickPlush}, $135$ demonstrations for \textit{CubeSort} and $200$ demonstrations for \textit{CoffeeMaking}.

We first test DP and KDPE for $50$ episodes on \textit{PickPlush} (Fig. \ref{Fig:real-tasks}), a picking task that consists of grasping and lifting a cat plush from the table. Additionally, mimicking the visual domain shift experiment performed in simulation, we test the same checkpoints on a version of \textit{PickPlush} where the plush is replaced with a yellow sponge, named \textit{PickSponge}. Then, we compare DP and KDPE for $100$ episodes on \textit{CubeSort}. The robot is required to grasp a blue and an orange cube from a red plate and place them in the cups of the corresponding colors. This requires the policy to learn a task with a high degree of multimodality both in the order in which the cubes are picked and in the way they are manipulated. In the task demonstrations, the two cubes are grasped in a random sequence and, during model evaluation, the cubes are randomly tossed on the plate at the beginning of each episode. Finally, we evaluate for $10$ episodes the two methods on the \textit {CoffeMaking} task, a real-world experiment to understand the impact of KDPE in a realistic scenario requiring long-horizon planning capabilities and considerable precision. The \textit{CoffeeMaking} task requires the robot to complete the following steps: picking up a coffee pod, inserting the pod in the coffee machine with poor visual conditions, pushing the front part of the coffee machine to close it, and pulling down the metal handle. Some of these steps require high precision, e.g. for pod insertion, as it can be noticed from Fig.~\ref{Fig:real-tasks}, where the task is autonomously performed by KDPE.

Following the methodology described in Section~\ref{ssec:benchmark-setup}, we train the CNN-based version of the DP model for $80k$ steps from images of size 480$\times$480 and we evaluate DP and KDPE on the same checkpoint, using DDIM~\citep{song2020ddim} as noise scheduler to speed-up inference.

\section{Results}

\begin{table}[t]
  \centering
  \small
  \setlength{\tabcolsep}{3pt}
  \begin{tabular}{>{\raggedleft\arraybackslash}p{1.95cm}|cc|cc|cc|c|c|c|c|c}
    \toprule
    & \multicolumn{2}{c|}{\textit{Lift}} & \multicolumn{2}{c|}{{Can}} & \multicolumn{2}{c|}{\textit{Square}} & \multicolumn{1}{c|}{\textit{ToolHang}} & \multicolumn{1}{c|}{\textit{Coffee}} & \multicolumn{1}{c|}{\textit{Stack}} & \multicolumn{1}{c|}{\textit{Assembly}} & Average\\
    & \textit{ph} & \textit{mh} & \textit{ph} & \textit{mh} & \textit{ph} & \textit{mh} & \textit{ph} &       & & &\\
    \midrule
    DP-C & \textbf{100}&\textbf{100}& 96.7& 94.0& \textbf{93.0}& 83.0& 62.0& 85.0& 80.7& 54.3& 84.9 \\
    KDPE-OOD-C& \textbf{100}&\textbf{ 100}& 96.0& 84.0& 83.0 &49.0& 7.0& 85.0& 72.0& 40.0&71.6 \\
    Tr-KDPE-C& \textbf{100}&\textbf{ 100}& 97.0& 94.0& 90.0&78.0& 68.0& 76.0& 78.0& \textbf{61.0}&84.2 \\
    KDPE-C & \textbf{100}& \textbf{100}&\textbf{ 98.0}& \textbf{96.0}& 92.0& \textbf{86.0}& \textbf{76.0}& \textbf{88.0}& \textbf{85.0}& \textbf{61.0}& \textbf{88.2} \\
    \midrule
    DP-T & \textbf{100}& \textbf{100}& 96.0& 90.0& 84.0& 73.0& 61.3&91.7& 73.7& \textbf{23.3}& 79.3 \\
    KDPE-OOD-T& \textbf{100}& 99.0& 95.0& 91.0& 84.0&65.0& 2.0& \textbf{93.0}& 61.0& 23.0&71.3 \\
    Tr-KDPE-T& \textbf{100}& 99.0& 93.0& 92.0& 81.0&81.0& 62.0& 86.0& \textbf{74.0}& 18.0&78.6 \\
    KDPE-T & \textbf{100}& \textbf{100}&\textbf{ 97.0}& \textbf{92.0}& \textbf{88.0}& \textbf{83.0}& \textbf{66.0}& 91.0& 72.0& 23.0& \textbf{81.2} \\
    \bottomrule
  \end{tabular}
  \caption{Average success rate (\%) of DP, KDPE-OOD, Tr-KDPE and KDPE on RoboMimic and MimicGen. The last column reports the average success rate over all the tasks.}
  \label{Tab:standard}
  \vspace{-0.4cm}
\end{table}

\begin{table}[t]
    \centering
    \small
    \setlength{\tabcolsep}{3pt} 
    \begin{tabular}{>{\raggedleft\arraybackslash}p{1.95cm}|cc|cc|cc|c|c|c|c|c}
    \toprule
    & \multicolumn{2}{c|}{\textit{Lift}} & \multicolumn{2}{c|}{\textit{Can}} & \multicolumn{2}{c|}{\textit{Square}} & \multicolumn{1}{c|}{\textit{ToolHang}} & \multicolumn{1}{c|}{\textit{Coffee}} & \multicolumn{1}{c|}{\textit{Stack}} & \multicolumn{1}{c|}{\textit{Assembly}} & Average\\
    & \textit{ph} & \textit{mh} & \textit{ph} & \textit{mh} & \textit{ph} & \textit{mh} & \textit{ph} &       & & &\\
        \midrule
        DP-C   & \textbf{100} & 99.6 & 92.3 & \textbf{91.3} & 93.7 & 79.7 & 59.0 & 92.3& 75.3& 44.7& 82.8 \\
         KDPE-C & \textbf{100} & \textbf{100} & \textbf{94.0} & 90.0 & \textbf{94.0} & \textbf{90.0} & \textbf{69.0} & \textbf{97.0}& \textbf{88.0} & \textbf{50.0} & \textbf{87.2} \\
        \midrule
        DP-T   &       98.7&       \textbf{100}&       89.0&       84.0&       \textbf{79.3}&       68.3&       58.0& \textbf{73.3}& 67.7& 15.6&       73.4 \\
        KDPE-T &       \textbf{99.0}&       \textbf{100}&       \textbf{90.0}&       \textbf{85.0}&       75.0&       \textbf{77.0}&       \textbf{62.0}& 72.0& \textbf{68.0}& \textbf{20.0}&       \textbf{74.8} \\
        \bottomrule
    \end{tabular}
    \caption{Success rate (\%) of DP and KDPE under object color perturbation.}
    \label{Tab:object_color}
    \vspace{-0.7cm}
\end{table}

\subsection{Benchmark on RoboMimic and MimicGen Tasks}
\label{sec:benchmark}

Results in the benchmark reported in Tab.~\ref{Tab:standard} show that KDPE outperforms DP on both architectures, and that KDPE-C is overall the best-performing method, achieving an average success rate of 88.2\%. KDPE improves DP by 3.3\% when using the CNN backbone, and by 1.9\% with the Transformer-based model. These gaps are even more pronounced (4.2\% and 2.4\%) if we exclude from the analysis the \textit{Lift} task which saturates to a success rate of 100\% for both methods. It is also worth noting that, in the three experiments where DP outperforms KDPE, the performance gap is limited and the success rate of KDPE is always within the standard deviation over the three trials of DP (see App.~\ref{app:std}). 

KDPE achieves a larger performance improvement w.r.t. DP on precision tasks as 
\textit{ToolHang}, and on tasks learned from lower quality data (\textit{mh}). The first observation indicates that the success rate of high-precision tasks is more influenced by outlier trajectories. The second, paired with the intuition that policies trained on noisy data learn to generate the outliers contained in the dataset, underlines the importance of our filtering mechanism. These findings are further supported by the performance of KDPE-OOD that, in a specular way, experiences its largest drops in performance on the tasks where KDPE performs best. The trajectory-based version of KDPE achieves similar performance to DP, resulting in no relevant improvement on either architecture. We attribute this to the fact that, to analyze the whole trajectory, Tr-KDPE needs a higher-dimensional kernel. This might lead to curse of dimensionality and poor characterization of the population's distribution. While the population size could be increased, it is important to notice that the required sample size increases exponentially with respect to dimensionality. This would lead to an exponential increase of the computational cost.

\subsection{Analysis Under Visual Environment Perturbations}
\label{sec:perturbation}

To study the robustness of DP and KDPE to domain shift, we perturb the model observations through the object color modification presented in Sec.~\ref{sec:expsetup_sim}, and measure the success rate on the same set of RoboMimic and MimicGen tasks considered in Sec.~\ref{sec:benchmark}. Results in Tab.~\ref{Tab:object_color} show that, on average, KDPE outperforms DP with both CNN and Transformer-based models by a similar margin to that presented in Tab.~\ref{Tab:standard} for the benchmark experiments in the unperturbed setting. This result further supports the effectiveness of KDPE in filtering out trajectory outliers, even in noisier settings. 

\subsection{Real Robot Results}

\begin{table}[t]
    \centering
    \small
    \setlength{\tabcolsep}{4pt} 
    \begin{tabular}{r|c|c|c|c}
        \toprule
        &\textit{PickPlush} &\textit{PickSponge} &\textit{CubeSort} & \textit{CoffeeMaking} \\
        \midrule
        DP-C   & 90 & 88 & 41 & 60 \\
        KDPE-C & \textbf{96} & \textbf{90} & \textbf{44} & \textbf{70} \\
        \bottomrule
    \end{tabular}
    \caption{Success rate (\%) of DP-C and KDPE-C on the real-world tasks. We test \textit{PickPlush} and \textit{PickSponge} for 50 episodes, \textit{CubeSort} for 100 episodes and \textit{CoffeeMaking} for 10 episodes.}
    \label{Tab:table_robot}
    \vspace{-0.6cm}
\end{table}

We compare KDPE to DP on three real-world tasks to study whether the improvement in performance observed in the simulated tasks translates to better performance on the real robot. Results in Tab.~\ref{Tab:table_robot} show that in the four experiments KDPE outperforms DP in terms of success rate. Moreover, for all the tasks, we noticed a smoother behavior of robot end-effector and gripper aperture. This suggests that KDPE allows to choose trajectories, representing task modalities, more consistently. 

\textbf{PickPlush} We evaluate performance for 50 episodes on the two different task configurations described in Sec.~\ref{ssec:robot-setup}. We initialize each episode with the plush in a random position. However, when one of the two methods fails to grasp the object, we perform a rollout of the other method starting from the same object position. KDPE solves the task with the orange plush, \textit{PickPlush}, 48 times, while DP 45 (96\% vs. 90\% success rate), being in line with the KDPE-C results in Tab.~\ref{Tab:standard}. In the three failure cases of DP, the choice of a suitable trajectory with KDPE has been critical to solve the task. We also test the same model (trained only to grasp the orange plush) with the yellow sponge (\textit{PickSponge}). Similarly to the experiments under object color perturbation presented in Sec.~\ref{sec:perturbation}, KDPE outperforms DP, but the gap in performance w.r.t. \textit{PickPlush} reduces. This may indicate that, on the failed episodes with the yellow sponge, DP (and therefore KDPE) could not predict suitable in-distribution trajectories. 

\textbf{CubeSort} We evaluate the two algorithms for 100 episodes, by randomly tossing the two cubes on the plate at the beginning of each episode. Differently from \textit{PickPlush}, we could not test the methods on the same object positions, as the failure state of the task is often partial (i.e., only one cube is positioned in the matching cup). KDPE achieves a higher success rate (3\% higher) which reflects the results observed in simulation for KDPE-C, e.g. on the \textit{Stack} task which requires to manipulate cubes of a similar size.

\textbf{CoffeeMaking} We report performance for the \textit{CoffeeMaking} task for 10 episodes. We noticed that, for this task, the initial position of the pod on the table has a huge impact on the final success of the task. Therefore, for a fair comparison, we tested both methods with the same pod initializations. KDPE managed to solve the task seven times, while DP six. 

\subsection{Inference Time}
We measure the inference time of DP-C, KDPE-C and Tr-KDPE-C with DDIM sampling on the machine used for the real-world experiments (equipped with an NVIDIA RTX 3080 GPU). DP-C, which in the original implementation generates a single trajectory, requires $77$ms, while generating a population of 100 trajectories takes an additional $13$ms, and performing KDE on them adds only $3$ms. Therefore, DP-C can be executed at a control frequency of $\sim12.99$Hz, while KDPE-C at $\sim10.75$Hz, meaning that KDPE-C adds a computational overhead of only $\sim2.24$Hz. For Tr-KDPE-C, instead, performing KDE requires 30ms, and can be executed at $\sim8.33$Hz.

\section{Conclusion}

DP has recently gained popularity as one of the most effective methods to train behavior cloning policies, thanks to its policy parameterization as a DDPM which allows to model the multimodality in the training data. While being effective, however, the sampling process is not constrained in any way. This can lead the model to predict trajectories that take the robot out-of-distribution. We propose to overcome this limitation with KDPE, a strategy to select with KDE the most representative trajectory computed by different DP denoising processes.

We quantitatively benchmarked KDPE against DP on four RoboMimic and three MimicGen simulated tasks, and on three real robot experiments: a tabletop plush picking, a multimodal and multi-step cube sorting, and a long-horizon coffe making task that requires high precision to be completed.

We showed that KDPE achieves better performance in the lower demonstration-quality regime and on tasks that require higher precision, also in presence of visual environment perturbations.

Two interesting directions for future research are using KDPE to guide the denoising process of DP, and applying KDPE to higher-dimensional problems, such as bimanual or dexterous manipulation tasks with anthropomorphic hands. However, such applications could incur the additional challenge of addressing the curse of dimensionality for KDE, a known challenge for kernel-based methods.

\section{Limitations}

\noindent{\textbf{Limiting assumptions}} KDPE selects certain output trajectories among the ones produced by DP. If this latter does not generate a representative distribution of trajectories, e.g. due to poor training, KDPE cannot provide any performance improvement. 

\noindent{\textbf{{Failure modes}} One of the strengths of DP is that the stochasticity of the trajectory generation can help the robot recover from out-of-distribution states by generating trajectories that may be associated to smaller probability density. KDPE aims to reduce the probability that the robot ends up in such states. However, detecting when this happens (e.g., from the estimated PDF from KDE) and switching to a different trajectory selection method would be an interesting improvement to KDPE.

\noindent{\textbf{Limitations of the results and experiments}} In this paper we proposed KDPE as a component that can be plugged on top of DP. However, it would be interesting to apply KDPE to other generative models like flow matching~\citep{zhang2023bootstrap}, or to generalist policies~\citep{octo_2023, intelligence2025pi_, kim24openvla}. 

\section*{Acknowledgements}
This research was supported by the PNRR MUR project PE0000013-FAIR, by the National Institute for Insurance against Accidents at Work (INAIL) project ergoCub-2.0, and by the Brain and Machines Flagship Programme of the Italian Institute of Technology.

\bibliography{bibliography}
\newpage
\appendix 

\section{Analysis Under Visual Environment Perturbations}
\label{app:perturbations}

In the experiments under visual environment perturbation presented in Sec.~\ref{sec:expsetup_sim}, we slightly modify the color of an object in each environment. Specifically, if the object has a texture  we compute its average color, we reduce its lightness (third channel under the HSL color scheme) by 10\% and set it as the object color. The object modified in the simulated task environments are shown in Fig.~\ref{Fig:shift}.

\begin{figure}[htbp]
    \centering
    \includegraphics[width=1.0\linewidth]{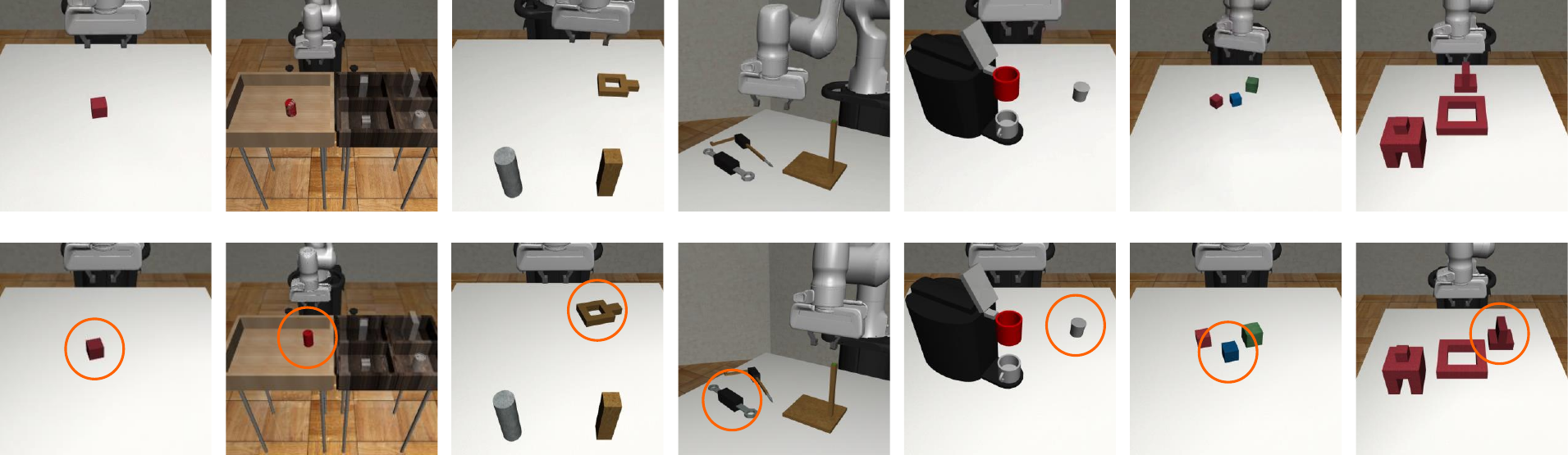}
    \caption{First row: unperturbed task environments. Second row: perturbed environments with the objects modified in the color perturbation experiments highlighted in the orange circles.}
    \label{Fig:shift}
\end{figure}

\section{Definition of Tr-KDPE}
\label{app:trkdpe}
 Tr-KDPE (Trajectory-KDPE) is a modification of the KDPE algorithm that estimates the probability density function (PDF) over a population of complete action trajectories. We recall that KDPE instead selects action trajectories based only on the density associated to the final action. Therefore, Tr-KDPE mitigates the risk of favoring outlier trajectories that, while ending in an in-distribution state for KDPE, may overall represent out-of-distribution trajectories. 

To model the joint PDF $p(\mathbf{a}_1, ..., \mathbf{a}_T)$ of the entire action sequence, Tr-KDPE uses the Markov assumption to decompose this joint probability into a product of conditional probabilities: $p(\mathbf{a}_1, ..., \mathbf{a}_T) = p(\mathbf{a}_1) \prod_{t=2}^{T} p(\mathbf{a}_t | \mathbf{a}_{t-1})$. The estimation relies on modeling the transition probabilities $p(\mathbf{a}_t | \mathbf{a}_{t-1})$ with conditional Kernel Density Estimation~\citep{hyndman_ckde_1996}:
 \begin{equation}
\hat{f}(y | x) = \frac{\hat{g}(x, y)}{\hat{h}(x)},
\end{equation}
where $\hat{g}(x, y)$ is the KDE of the joint density $g(x, y)$
\begin{equation}
\hat{g}(x, y) = \frac{1}{N} \sum_{j=1}^{N} k\left(x ,x_j\right) k\left(y ,y_j\right)
\end{equation}
and $\hat{h}(x)$ is the KDE of the marginal density $h(x)$
\begin{equation}
\hat{h}(x) = \frac{1}{N} \sum_{j=1}^{N}k\left(x ,x_j\right).
\end{equation}
Therefore, we can compute the kernel density estimator of $p(\mathbf{a}_t | \mathbf{a}_{t-1})$ as:
 \begin{equation}
\rho(\mathbf{a}_t | \mathbf{a}_{t-1}) = \frac{\hat{g}(\mathbf{a}_{t-1}, \mathbf{a}_t)}{\hat{h}(\mathbf{a}_{t-1})}.
\end{equation}
Consequently, estimating the full trajectory density $p(\mathbf{a}_1, ..., \mathbf{a}_T)$ involves computing one standard KDE for $p(\mathbf{a}_1)$ and $T-1$ conditional KDEs for the terms $t=2, ..., T$.

\section{Hyperparameters}
\label{app:params}

The hyperparameters used during training and evaluation are reported in Tab.~\ref{Tab:hyperparameters}. 

\begin{table}[!htbp]
\centering
\label{tab:hyperparameters_modified}
\begin{tabular}{@{}c|c@{}}
\toprule
\textit{Hyperparameter}                     & \textit{Value}                                         \\
\midrule
Optimizer                          & Adam                                          \\
Learning Rate                      & $1 \times 10^{-4}$                            \\
Betas                              & $(0.95, 0.999)$                               \\
Epsilon                            & $1 \times 10^{-8}$                            \\
Weight Decay                       & $1 \times 10^{-6}$                            \\
Noise scheduler                          & DDPM (simulation) / DDIM (real robot)       \\
Diffusion Inference Steps          & 100 (simulation) / 10 (real robot)          \\
Total Training Steps               & 80,000                                        \\
Population Size ($N$)                & 100                                           \\
Execution Horizon ($T$)              & 8                                             \\
Action Dimensionality ($D$)          & 10                                            \\
Position Bandwidth ($\sigma_{pos}$)  & 0.05                                          \\
Rotation Bandwidth ($\sigma_{rot}$)  & 0.25                                          \\
Gripper Bandwidth ($\sigma_{grip}$)  & 1.0                                           \\
\bottomrule
\end{tabular}
\caption{Hyperparameters used for training and evaluation of KDPE. Symbols in the parentheses correspond to those used in Sec.~\ref{sec:methodology}.}
\label{Tab:hyperparameters}
\end{table}

\section{DP Standard Deviations}
\label{app:std}

As reported in Sec.~\ref{sec:kdpe}, since the baseline performance (DP-C or DP-T) depends on the random sampling of a trajectory, we repeated its evaluation three times with three different seeds. We report the performance of all the methods on the benchmark experiments with DP standard deviations in Tab.~\ref{Tab:benchmark_stdev}, and on the experiments under object color perturbation in Tab.~\ref{Tab:pert_stdev}. 

\begin{table}[htbp]
  \centering
  \small
  \setlength{\tabcolsep}{3pt}

  \begin{tabular}{@{}r|cc|cc|cc|c@{}}
    \toprule
    & \multicolumn{2}{c|}{\textit{Lift}} & \multicolumn{2}{c|}{{Can}} & \multicolumn{2}{c|}{\textit{Square}} & \multicolumn{1}{c} {\textit{ToolHang}} \\ 
     & \textit{ph} & \textit{mh} & \textit{ph} & \textit{mh} & \textit{ph} & \textit{mh} & \textit{ph} \\
    \midrule
    DP-C & $\boldsymbol{100 \pm 0}$ & $\boldsymbol{100 \pm 0}$ & $96.7 \pm 1.00$ & $94.0 \pm 2.00$ & $\boldsymbol{93.0 \pm 3.00}$ & $83.0 \pm 7.00$ & $62.0 \pm 2.00$\\
    KDPE-OOD-C & $\boldsymbol{100}$ & $\boldsymbol{100}$ & $96.0$ & $84.0$ & $83.0$ & $49.0$ &  $7.00$ \\
    Tr-KDPE-C & $\boldsymbol{100}$ & $\boldsymbol{100}$ & $97.0$ & $94.0$ & $90.0$ & $78.0$ & $68.0$\\
    KDPE-C & $\boldsymbol{100}$ & $\boldsymbol{100}$ & $\boldsymbol{98.0}$ & $\boldsymbol{96.0}$ & $92.0$ & $\boldsymbol{86.0}$ & $\boldsymbol{76.0}$\\
    \midrule
    DP-T & $\boldsymbol{100 \pm 0}$ & $\boldsymbol{100 \pm 0}$ & $96.0 \pm 2.00$ & $90.0 \pm 3.00$ & $84.0 \pm 4.00$ & $73.0 \pm 7.00$ & $61.3 \pm 10.0$\\
    KDPE-OOD-T & $\boldsymbol{100}$ & $99.0$ & $95.0$ & $91.0$ & $84.0$ & $65.0$ & $2.00$ \\
    Tr-KDPE-T & $\boldsymbol{100}$ & $99.0$ & $93.0$ & $92.0$ & $81.0$ & $81.0$ &  $62.0$ \\
    KDPE-T & $\boldsymbol{100}$ & $\boldsymbol{100}$ & $\boldsymbol{97.0}$ & $\boldsymbol{92.0}$ & $\boldsymbol{88.0}$ & $\boldsymbol{83.0}$ & $\boldsymbol{66.0}$\\
    \bottomrule
  \end{tabular}

  \vspace{\medskipamount} 
  \begin{tabular}{@{}r|c|c|c@{}}
    \toprule
     & \multicolumn{1}{c|}{\textit{Coffee}} & \multicolumn{1}{c|}{\textit{Stack}} & \multicolumn{1}{c}{\textit{Assembly}}\\
    \midrule
    DP-C & $85.0 \pm 6.00$ & $80.7 \pm 1.00$ & $54.3 \pm 1.00$\\
    KDPE-OOD-C & $85.0$ & $72.0$ & $40.0$\\
    Tr-KDPE-C & $76.0$ & $78.0$ & $\boldsymbol{61.0}$\\
    KDPE-C & $\boldsymbol{88.0}$ & $\boldsymbol{85.0}$ & $\boldsymbol{61.0}$\\
    \midrule
    DP-T & $91.7 \pm 6.00$ & $73.7 \pm 2.00$ & $\boldsymbol{23.3 \pm 2.00}$\\
    KDPE-OOD-T & $\boldsymbol{93.0}$ & $61.0$ & $23.0$\\
    Tr-KDPE-T & $86.0$ & $\boldsymbol{74.0}$ & $18.0$\\
    KDPE-T & $91.0$ & $72.0$ & $23.0$\\
    \bottomrule
  \end{tabular}

  \caption{Benchmark results presented in Tab.~\ref{Tab:standard}, reporting also the standard deviation of the success rate (\%) over three runs of DP. Top table: RoboMimic (\textit{Lift}, \textit{Can}, \textit{Square}, \textit{ToolHang}). Bottom table: MimicGen (\textit{Coffee}, \textit{Stack}, \textit{Assembly}).}
  \label{Tab:benchmark_stdev}
\end{table}

\begin{table}[htbp]
  \centering
  \small
  \setlength{\tabcolsep}{3pt}

  \begin{tabular}{@{}r|cc|cc|cc|c@{}}
    \toprule
    & \multicolumn{2}{c|}{\textit{Lift}} & \multicolumn{2}{c|}{{Can}} & \multicolumn{2}{c|}{\textit{Square}} & \multicolumn{1}{c} {\textit{ToolHang}} \\ 
     & \textit{ph} & \textit{mh} & \textit{ph} & \textit{mh} & \textit{ph} & \textit{mh} & \textit{ph} \\
    \midrule
    DP-C   & $\boldsymbol{100 \pm 0}$ & $99.6 \pm 0.57$ & $92.3\pm4.16$ & $\boldsymbol{91.3 \pm 3.21}$ & $93.7\pm2.52$ & $79.7\pm4.16$ & $59.0\pm1.53$ \\
    KDPE-C & $\boldsymbol{100}$ & $\boldsymbol{100}$ & $\boldsymbol{94.0}$ & $90.0$ & $\boldsymbol{94.0}$ & $\boldsymbol{90.0}$ & $\boldsymbol{69.0}$ \\
    \midrule
    DP-T   &       $98.7\pm0.58$&       $\boldsymbol{100 \pm 0}$&       $89.0\pm4.58$&       $84.0\pm3.61$&      $\boldsymbol{79.3\pm1.53}$&       $68.3\pm2.51$&   $58.0\pm5.57$\\
    KDPE-T &       $\boldsymbol{99.0}$&       $\boldsymbol{100}$&       $\boldsymbol{90.0}$&       $\boldsymbol{85.0}$&       75.0&       $\boldsymbol{77.0}$&       $\boldsymbol{62.0}$\\
    \bottomrule
  \end{tabular}

  \vspace{\medskipamount} 
  \begin{tabular}{@{}r|c|c|c@{}}
    \toprule
     & \multicolumn{1}{c|}{\textit{Coffee}} & \multicolumn{1}{c|}{\textit{Stack}} & \multicolumn{1}{c}{\textit{Assembly}}\\
    \midrule
    DP-C & $92.3\pm5.03$& $75.3\pm0.58$& $44.7\pm1.50$\\
    KDPE-C & $\boldsymbol{97.0}$& $\boldsymbol{88.0}$ & $\boldsymbol{50.0}$\\
    \midrule
    DP-T   & $\boldsymbol{73.3\pm7.37}$& $67.7\pm4.16$ &$15.6\pm4.16$\\
    KDPE-T & $72.0$& $\boldsymbol{68.0}$& $\boldsymbol{20.0}$\\
    \bottomrule
  \end{tabular}

  \caption{Results of experiments under object color perturbation presented in Tab.~\ref{Tab:object_color}, reporting also the standard deviation of the success rate (\%) over three runs of DP. Top table: RoboMimic (\textit{Lift}, \textit{Can}, \textit{Square}, \textit{ToolHang}). Bottom table: MimicGen (\textit{Coffee}, \textit{Stack}, \textit{Assembly}).}
  \label{Tab:pert_stdev}
\end{table}

\newpage
\section{Visualizer}
During the development and study of KDPE and comparative baselines, we developed a visualizer using the \textit{rerun} data visualization library~\citep{RerunSDK}. This tool facilitates the analysis of populations of trajectories, including action positions, orientations, and gripper states. Through the help of the visualizer, we studied how the KDE bandwidths (reported in the last three rows of Tab.~\ref{Tab:hyperparameters}) influence the action densities, and found suitable values that let KDE capture DP output multimodality. Fig.~\ref{Fig:visualizer} shows a snapshot of the visualizer.

We open-source the code of the visualizer, hoping it will assist others in the analysis of generative robotic policies. The code for the visualizer is available on the project page at \url{https://hsp-iit.github.io/KDPE/}.

\begin{figure}[htbp]
    \centering
    \includegraphics[width=0.5\linewidth]{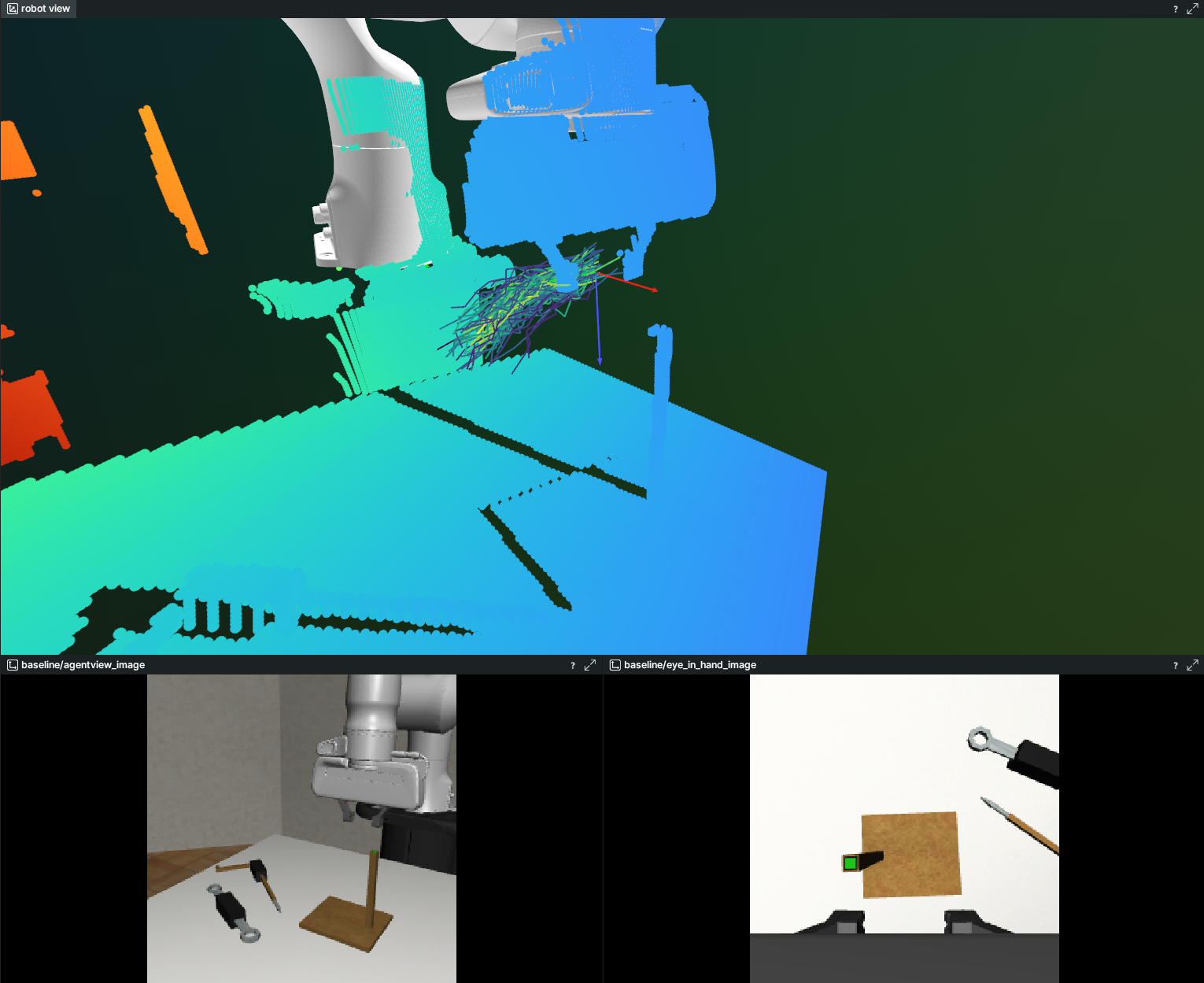}
    \caption{The trajectory visualizer being used to analyze trajectories on the RoboMimic \textit{ToolHang} task. The scene in the 3D view (\textit{robot view} window) is represented as a point-cloud, since object meshes are not readily available for real-world environments. The visualizer supports assigning different colormaps to the population of trajectories. The colormap in the picture represents the densities assigned by KDPE to each trajectory.}
    \label{Fig:visualizer}
\end{figure}

\end{document}